\documentclass[10pt,twocolumn,letterpaper]{article}

\usepackage{iccv}
\usepackage{times}
\usepackage{epsfig}
\usepackage{graphicx}
\usepackage{amsmath}
\usepackage{amssymb}
\usepackage{float}

\usepackage{booktabs}
\usepackage{multirow}

\usepackage{url}
\usepackage{algorithm}
\usepackage{algorithmic}
\usepackage{enumitem}
\usepackage[table]{xcolor}

\renewcommand{\algorithmicrequire}{\textbf{Input:}}
\renewcommand{\algorithmicensure}{\textbf{Output:}}

\usepackage{caption}
\usepackage[breaklinks=true,bookmarks=false]{hyperref}

\iccvfinalcopy 

\begin{document}

\title{Lidar Light Scattering Augmentation (LISA): Physics-based Simulation of  Adverse Weather Conditions for 3D Object Detection}

\author{Velat Kilic, Deepti Hegde, Vishwanath Sindagi, A. Brinton Cooper, Mark A. Foster and Vishal M. Patel\\
Department of Electrical and Computer Engineering, Johns Hopkins University\\
Baltimore, U.S.A
}

\maketitle
\ificcvfinal\thispagestyle{empty}\fi

\begin{abstract}
Lidar-based object detectors are critical parts of the 3D perception pipeline in autonomous navigation systems such as self-driving cars. However, they are known to be sensitive to adverse weather conditions such as rain, snow and fog due to reduced signal-to-noise ratio (SNR) and signal-to-background ratio (SBR). As a result, lidar-based object detectors trained on data captured in normal weather tend to perform poorly in such scenarios. However, collecting and labelling sufficient training data in a diverse range of adverse weather conditions is laborious and prohibitively expensive. To address this issue, we propose a physics-based approach to simulate lidar point clouds of scenes in adverse weather conditions. These augmented datasets can then be used to train lidar-based detectors to improve their all-weather reliability. Specifically, we introduce a hybrid Monte-Carlo based approach that treats (i) the effects of large particles by placing them randomly and comparing their back reflected power against the target, and (ii) attenuation effects on average through calculation of scattering efficiencies from the Mie theory and particle size distributions. Retraining networks with this augmented data improves mean average precision evaluated on real world rainy scenes and we observe greater improvement in performance with our model relative to existing models from the literature. Furthermore, we evaluate recent state-of-the-art detectors on the simulated weather conditions and present an in-depth analysis of their performance.

\end{abstract}

\section{Introduction}

Autonomous navigation systems like self-driving cars have received increased interest in the recent years. As a result, 3D object detection, which forms a critical backbone in these systems, has witnessed rapid progress. Several works such as \cite{pang2020clocs,liang2019multi,ku2019monocular,yang20203dssd,yang2018pixor,Shi2019PartA2N3,zhou2018voxelnet} have been successful in improving the performance on different benchmark datasets. A majority of these approaches \cite{zhou2018voxelnet,yan2018second,lang2019pointpillars,yang20203dssd,shi2020pv} are based on lidar data, primarily due to the fact that these sensors provide highly accurate range measurements which enable reliable detection performance. 

However, these sensors are known to be sensitive to adverse weather conditions \cite{bmw,goodin,prin,Heinzler}  such as rain, snow and fog due to reduced SNR and SBR as well as large back scattered power from random droplets. As the laser pulse traverses the scattering medium, the total intensity is attenuated exponentially with distance due to scattering. This results in weather-dependent changes in reflectivities and increased range uncertainty. In some cases, this can cause mis-detections if the signal is reduced below the noise floor. Further, the background strength is increased especially close to the sensor simply because a larger fraction of the back scattered laser power from random droplets is available to the sensor. Especially in the strongest return mode, this could confuse the detector as to which pulse is coming from the real target and result in randomly scattered points close to the sensor and correspondingly missing points within the real target. All these effects can be visually observed from lidar scans shown in Figs \ref{fig:lidar_compare} and \ref{fig:viz} and of course they in turn adversely affect the performance of lidar-based detectors that are typically trained on clean datasets\footnote{Datasets that do not contain samples affected by adverse weather.}. To address this issue, one approach would be to collect data such that explicit attempt is made to capture these adverse conditions. However, this process can be labour intensive and prohibitively expensive. Additionally, it would not be possible to obtain identical lidar scenes in normal weather, which is often required in data-driven methods for 3D scene understanding. Hence, in this work, we focus on the task of simulating the adverse weather conditions on existing lidar-based datasets. Our model will enable easy generation of realistic lidar point clouds for a variety of weather conditions, which can be used to study the effects of weather on 3D object detectors in addition to improving their overall all-weather reliability.

Notably, there are only a few works \cite{bmw,goodin,prin,Heinzler} that attempt to tackle this challenge. For example, Rasshofer \etal \cite{bmw} theoretically model the average effects of rain, fog and snow through calculation of the extinction coefficient and the geometry of bi-static lidars. Recently, Bijelic \etal \cite{prin} modeled foggy conditions by assuming that the emitted laser beam intensity is attenuated by a transmission factor and introducing randomly scattered points. Goodin \etal \cite{goodin} develop a more physical simulator by modelling rain as a function of rain-rate and introducing rain rate dependent range uncertainty.

While these works attempt to develop simplified models for simulating different weather conditions, they suffer from the following drawbacks: (i)  In \cite{prin}, the positions of the randomly scattered points are selected from a uniform random distribution. This is a gross oversimplification since detected laser power is a strong function of range. In other words, rain droplets very close to the sensor are more likely to back scatter enough laser power to confuse the detector (see blue box in Fig \ref{fig:viz}), (ii) In \cite{prin, Heinzler}, the signal-to-noise ratio (SNR) has an additive gain factor: $SNR = \frac{L+g}{n}$ where $L$ is laser intensity, $g$ is gain and $n$ is noise which is clearly not physical since gain always multiplies SNR, \ie, $\frac{gL}{n}$. The SNR term is then used to predict which points should be randomly scattered and this error biases the reflectivities of the surviving points. In the extreme case, a signal with zero return ($L=0$) could still be reliably detected according to this incorrect model for large $g$ or small $n$ which is not physical. (iii) Further, the SNR dependent range uncertainty (as SNR decreases, range uncertainty increases) is typically ignored in the current models. This is very much related to the edge detection problem \cite{canny}; in the presence of noise, aggressive filtering is needed which then reduces localization accuracy. (iv) The extinction coefficients of different weather conditions are selected from a periodic distribution in \cite{prin} and an unknown distribution in \cite{Heinzler} within the bounds [$.001, 0.1$] $m^{-1}$. To overcome these drawbacks, we propose a hybrid Monte Carlo-based approach which considers physical atmospheric and lidar hardware parameters such as max and min (for bistatic configuration) lidar range, range accuracy, laser wavelength, rain rate, \etc. We use the proposed simulation model and perform an in depth analysis of the effect of adverse weather conditions on recent state-of-the-art lidar-based object detectors\cite{lang2019pointpillars,yang20203dssd,shi2020pv,he2020structure}.   \\

\noindent To summarize, the following are the main contributions of our work:
\begin{itemize}[topsep=0pt,noitemsep,leftmargin=*]
	\item We develop a physics-based simulation model to generate lidar point clouds for different adverse weather conditions.  Specifically, we propose a hybrid Monte Carlo based lidar scatterer simulator for rain, snow and different types of fog.
	\item We employ the proposed model to augment scans from the Waymo dataset and retrained the network on this augmented data. Using our model, we observe greater performance improvement on real world scenes with rain compared to the existing models.
	\item We employ the proposed model to analyze the performance degradation of 3D object detectors caused by weather, and compare it against the performance degradation caused by other simulation methods. 
\end{itemize}

\section {Related work}
\noindent\textbf{Simulation of weather effects}: Effects of weather related noise in lidar return signals have been investigated in the literature \cite{bmw,goodin} and weather models have been used to improve the performance of DNN-based denoisers and object detectors \cite{Heinzler,prin}. Rasshofer \etal provide physical models for studying the average effects of rain, fog and snow and study their influence on lidar sensors theoretically \cite{bmw}. \\
 
\noindent\textbf{3D object detection}:
Recent research on deep learning for point sets \cite{REF:qi2017pointnet,REF:qi2017pointnetplusplus} has enabled the end-to-end training of neural networks that can consume point cloud data without transforming them to intermediate representations, such as depth or bird's eye view formats. Qi \etal \cite{REF:qi2017pointnet} propose PointNet, a network that directly consumes point clouds and outputs class labels. Several methods offer multi-modal approaches to solve 3D object detection \cite{chen2017multi,liang2019multi,qi2018frustum}, with a combination of radar, lidar, and image data. In this work, we will focus on pure lidar methods when evaluating performance of detectors in adverse weather.

Methods which use deep neural networks for 3D object detection in lidar point clouds can be broadly categorized as voxel-based \cite{REF:zhou2017voxelnet,lang2019pointpillars,yan2018second}, point-based \cite{yang20203dssd,he2020structure,shi2019pointrcnn,qi2018frustum,ipod}, or as a combination of the two \cite{shi2020pv,yang2019std}. 

Zhou and Tuzel extend the PointNet framework to the 3D detection task by proposing VoxelNet \cite{REF:zhou2017voxelnet}, which involves the encoding of evenly sized voxels using feature embedding layers. Lang \etal propose PointPillars \cite{lang2019pointpillars}, which organizes points into voxels in the form of evenly spaced vertical columns which are encoded using a feature encoding block. 
Shi \etal propose PointRCNN \cite{shi2019pointrcnn} that involves point-wise feature extraction, 3D proposal generation and region-of-interest pooling based on \cite{REF:qi2017pointnetplusplus}. This method outperforms voxel-based approaches, but is computationally expensive. In \cite{yang20203dssd}, Yang \etal propose 3DSSD, a single stage, point-based object detection network which uses a fusion sampling method to eliminate the feature propagation layer used in similar networks \cite{REF:qi2017pointnetplusplus,shi2019pointrcnn} to improve the inference speed. In PV-RCNN \cite{shi2020pv}, Shi \etal integrate set abstraction from point-based methods and 3D convolutions from voxel CNNs by summarizing voxels into keypoints through a voxel set abstraction module. This method shows improvement over pure voxel-based \cite{lang2019pointpillars} and pure point based methods \cite{yang20203dssd}. This is further extended by Bhattacharyya \etal \cite{Bhattacharyya2020DeformablePI}, where they include an adaptive deformation and context fusion module for performing keypoint alignment.
 
\section{Preliminaries}
\begin{figure}[h]
    \centering
    \includegraphics[width=\linewidth]{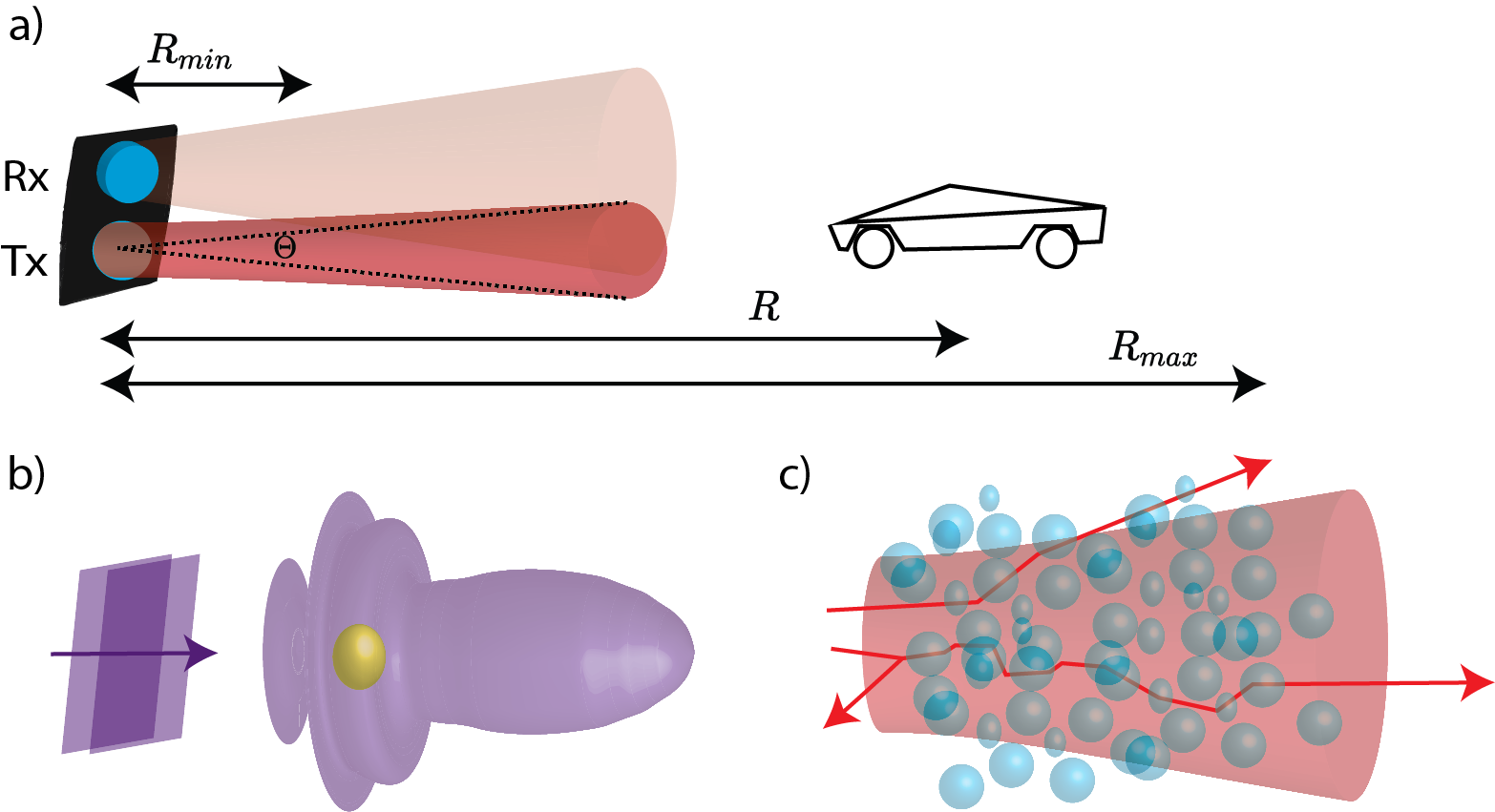}
    \caption{a) Bistatic lidar (where receiver-Rx and transmitter-Tx share different optical paths) parameters relevant for weather augmentation calculations. b) Light scattering by a single particle in the wave picture with a cartoon radiation pattern. c) Light scattering by a random ensemble of droplets in the ray picture.}
    \label{fig:lidar}
\end{figure}
As shown in Fig. \ref{fig:lidar}a, lidars measure target distances by sending pulses of light and measuring time of flight for the back-reflection. Important parameters for characterizing their weather performance are minimum range $R_{min}$, beam divergence $\Theta$ and maximum range $R_{max}$. Laser intensity is attenuated when traversing scattering media due to scattering and absorption as shown in Fig \ref{fig:lidar}c in the ray picture. The transmitted intensity $I_T$ through a scattering medium can be calculated using the Beer-Lambert law \cite{bmw}
\setlength{\belowdisplayskip}{0pt} \setlength{\belowdisplayshortskip}{0pt}
\setlength{\abovedisplayskip}{0pt} \setlength{\abovedisplayshortskip}{0pt}
\begin{equation}
    I_T = I_0 e^{- \int_0^R \alpha(r) dr},
    \label{beer}
\end{equation}
where $I_0$ is the incident intensity, $\alpha$ is the extinction coefficient and $R$ is the target range. The average extinction coefficient can be calculated from the extinction cross-section $\sigma_{ext}$ and the corresponding density as follows:
\begin{align}
    \alpha &= \int_0^{\infty} \sigma_{ext}(D) N(D) dD, \\
           &= \frac{\pi}{4} \int_0^{\infty} D^2 Q_{ext}(D) N(D) dD, \label{eq:alpha}
\end{align}
where $D$ is the particle diameter and $Q_{ext}$ is the extinction efficiency. Scattering cross-sections (in units of area i.e $m^2$) can be calculated for each particle geometry and material by solving Maxwell's equations and comparing total radiated flux ($W$) to incoming flux $W/m^2$ in the wave picture as shown in Fig.\ref{fig:lidar}b for a plane wave \cite{bohren}. Particle density function $N(D)$ is cited in units of $m^{-3} \, mm^{-1}$ (per volume $m^3$ per diameter $mm$) and modeled by the Marshall-Palmer distribution or gamma function models \cite{bmw}.

\begin{figure}[H]
\begin{center}
 \includegraphics[width=\linewidth]{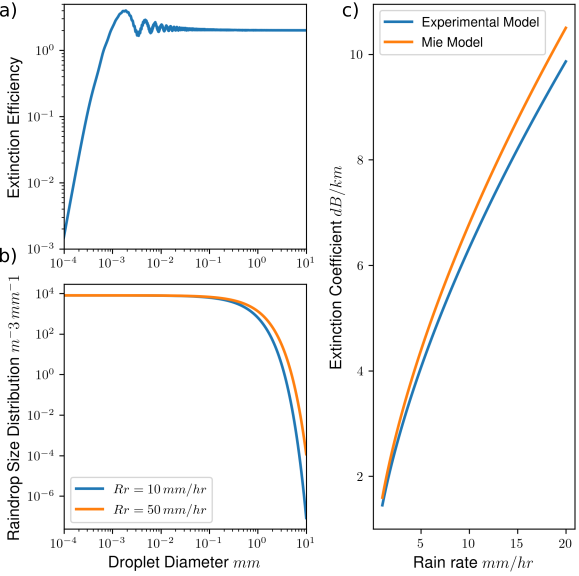}
 \caption{a) Extinction efficiency as a function of rain droplet diameter. b) Rain particle size distribution using the Marshall-Palmer distribution for different rain rates. c) Extinction coefficient as a function of rain rate as calculated numerically from Mie theory (orange line) and the asymptotic solution from \cite{ulbrich} (blue line) which is found to predict extinction coefficients within $25\%$ of the experimentally obtained values.}  \label{fig:rain}
 \end{center}
\end{figure}

Fig.~\ref{fig:rain}b illustrates the distribution of rain particle sizes as predicted by the Marshall-Palmer distribution for different rain rates on a logarithmic scale.

The next step is to calculate how a given extinction coefficient affects lidar return. Both the transmitted and back reflected pulses decay exponentially as they traverse scattering media which leads to a total attenuation of the back reflected signal by $e^{-2 \alpha R}$. Models in the literature reduce object reflectivities by that amount \cite{Heinzler,prin}. However, for exact values of these extinction coefficients, they sample probability distribution functions which fit measurements from a climate chamber. Although this is perfectly valid, we note that calculating extinction coefficients through physics would help model generalize to a greater number of atmospheric phenomena.
\\
Further, Bijelic \etal make the observation that if the SNR gets too small, then the lidar cannot reliably measure distance. Therefore, a maximum detectable range is defined
\begin{align}
    SNR &= \frac{P_r}{P_n}, \\
    R_{max} &= -\frac{ln\left( SNR^{-1} \right)}{2 \alpha},
\end{align}
where $P_r$ is the reflected power, $P_n$ is the noise power and $R_{max}$ is the max detectable range. However, SNR is estimated through the following unphysical function $SNR = \frac{L+g}{n}$, where the noise ($n$) and gain ($g$) parameters are fitted to experimental fog data \cite{Heinzler,prin}. In contrast, Goodin \etal model the detected power by considering both target reflectivity and range. Range corrected power is then compared against a minimum detectable power which is estimated from maximum lidar range \cite{goodin}.

\section{Proposed method}
Here we build on these previous models and extend them through physics-based arguments and by introducing a hybrid Monte-Carlo based approach. The algorithm takes in real design parameters such as maximum and minimum lidar range, range accuracy, laser beam divergence as well as real-world parameters such as rain rate. In order to map atmospheric conditions to optical parameters, we explicitly calculate the extinction coefficients from scattering theory. Extinction efficiency is a function of scatterer geometry and material composition and often does not have a closed form analytical solution \cite{bohren}. Therefore, we assume spherical particles for scatterers which can be calculated exactly from Mie theory. Although Mie theory provides an analytical solution, it requires summing an infinite series. Stopping the series summation at the correct point has been somewhat of an art \cite{bohren} and so in this work, we use an open source implementation, PyMieScatt \cite{pymie}, which has been extensively utilized and verified. Fig.~\ref{fig:rain}a shows the extinction efficiency as a function of rain droplet diameter for a 905 nm laser. Extinction efficiency along with particle distribution can be used to calculate an average extinction coefficient by numerically integrating the expression in Eq.~\ref{eq:alpha}. In order to test the accuracy of the Mie model against real weather phenomena, we compare it against an asymptotic relation which has been found to predict the extinction coefficients within $25\%$ of experimentally measured values for rain \cite{ulbrich}:
\begin{equation}
    \alpha = 1.45 Rr^{0.64},
\end{equation}
where $\alpha$ is in units of $dB/km$ and $Rr$ is the rain rate measured in $mm/hr$. In Fig.~\ref{fig:rain}c we compare the asymptotic model against our the Mie model using Marshall-Palmer rain droplet distribution. Although a simple expression might exist for a particular weather condition, we propose a model where effects of arbitrary weather phenomena can be computed efficiently. All the arguments up to this point are well suited for modelling effects of small and dense scatterers such as fog. However, strong scatterers such as large rain droplets, dust particles or snow flakes could reflect back more power than the target. In fact, a group of particles within the laser pulse volume (determined by pulse width and duration) could confuse the lidar detector even though each individual particle may not back scatter more power than the target.

\begin{algorithm}[H]
    \caption{LISA hybrid Monte-Carlo approach} 
    \label{alg:lisa}
    \algorithmicrequire{Lidar point cloud (pc), rain rate (Rr)} \\
    \algorithmicensure{Noisy point cloud (pcnew)}
    
    \begin{algorithmic}
    \STATE {Init pcnew}
        \FOR{x,y,z,$\rho$ (reflectivity) in pc}
            \STATE {Range $R = \sqrt{x^2 + y^2 + z^2}$}
            \STATE {Min power in arbitrary units $P_{min} = 0.9 R_{max}^{-2}$} 
            \STATE {Beam diameter $Db = R \, tan(\Theta)$ }
            \STATE {Density of droplets $Nd = N(D,Rr)$}
            \STATE {Extinction coefficient $\alpha$ from Eq.~\ref{eq:alpha}}
            \STATE{Nt = 0}
            \IF{$R>R_{min}$}
                \STATE {Conic beam volume $bvol = \frac{\pi}{3} R \left( \frac{Db}{2} \right)^2$}
                \STATE {Total number of particles above a certain diameter $D_s$ in beam path $Nt = bvol \, \int_{D_s}^{\infty} N(D) dD$}
                \STATE {Convert $Nt$ to integer with probabilistic rounding}
            \ENDIF
            \STATE {Randomly sample Nt many scatterers:
            \begin{itemize}[noitemsep,nolistsep]
                \item ranges ($R_{rand}$) from a quadratic PDF due to diverging beam \\
                \item reflectivities ($\rho_{rand}$) from Fresnel \\
                \item diameters ($D_{rand}$) from PDF associated with $N(D,Rr)$ \\
            \end{itemize}
            }
            \STATE {Discard points with range $R_{rand} < R_{min}$}
            \STATE {Back reflected power from the object  $P_0 = \frac{\rho e^{-2 \alpha R}}{R^2}$}
            \STATE {Back reflected power from random scatterers\\
            $P_r = \frac{\rho_{rand} e^{-2 \alpha R_{rand}}}{R_{rand}^2} \, min\left\{ \left( \frac{D_{rand}}{Db(R_{rand})} \right)^2, 1 \right\}$}
            \STATE {Pick random scatterer with max back reflected power $ind = argmax(P_{rand})$}
            \IF{$P_{0} < P_{min}$ and $P_{rand}[ind] < P_{min}$}
                \STATE {$R_{new} = 0$ and $\rho_{new} = 0$}
                \STATE {label = 0 for lost point}
            \ELSIF{$P_{rand}[ind] > P_0$}
                \STATE {$R_{new} = R_{rand}[ind]$ }
                \STATE {$\rho_{new} = \rho_{rand} e^{-2 \alpha R_{new}} \, min\left\{ \left( \frac{D_{new}}{Db(R_{new})} \right)^2, 1 \right\}$} 
                \STATE {label = 1 for randomly scattered point}
            \ELSE
                \STATE {Return original point with modified range and reflectivity:}
                \STATE {Range uncertainty $\sigma_R$from Eq.~\ref{eq:ran}}
                \STATE {New range $R_{new} = R + \mathcal{N}(0,\sigma_R)$}
                \STATE {New reflectivity $\rho_{new} = \rho e^{-2 \alpha R}$}
                \STATE {label = 2 for original point}
            \ENDIF
            \STATE {$\theta = cos^{-1}\left( z/R \right)$, $\phi = arctan2(y,x)$}
            \STATE {$x_{new} = R_{new} sin(\theta) cos(\phi)$}
            \STATE {$y_{new} = R_{new} sin(\theta) sin(\phi)$}
            \STATE {$z_{new} = R_{new} cos(\theta)$}
            \STATE {pcnew.append$ \left( x_{new}, y_{new}, z_{new}, \rho_{new} \right)$}
        \ENDFOR
    \end{algorithmic}
\end{algorithm}

For our algorithm, we ignore the latter effect to keep computations simple which is equivalent to assuming infinitesimally small pulse width. Range uncertainty arising from the finite pulse width is already taken into account by Eq.~\ref{eq:ran}. The only effect missed by this assumption is scattering from non-homogeneous clouds of particles which could be manually added later. A full Monte-Carlo approach would randomly place all scatterers, calculate the back reflected power from each particle and then convolve that with the pulse shape. 

However, in order to keep computations simple, we only consider particles above a certain diameter since small particles are not very likely to  back scatter enough light to confuse the detector.
\begin{figure}[H]
    \centering
    \includegraphics[width=\linewidth]{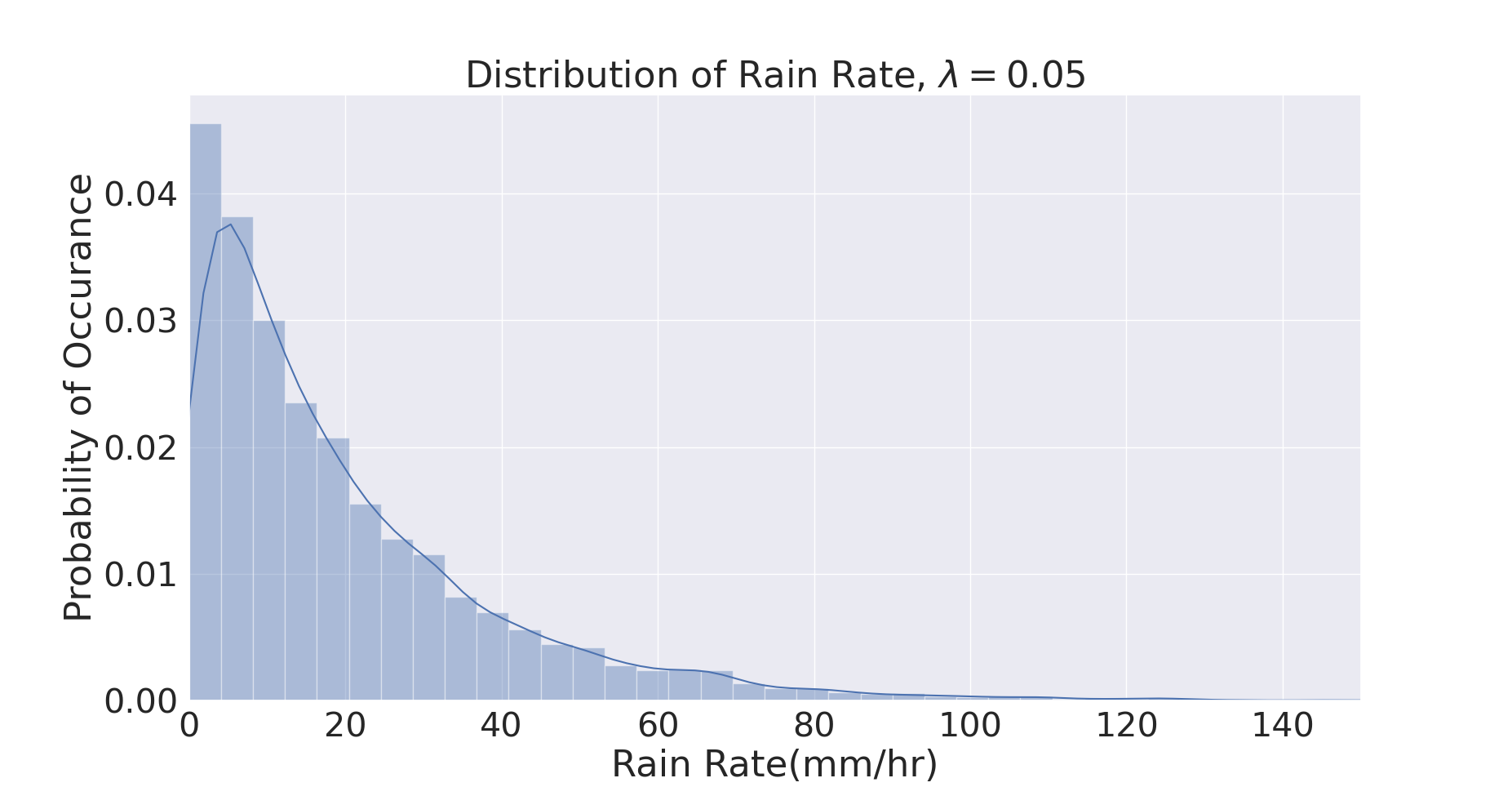}
   
    \caption{Graph of probability of occurrence of rain rate versus rain rate in $mm/hr$, showing the distribution of rain rates across lidar scenes in simulated adverse weather conditions. Rain rates are sampled from this distribution for all simulation experiments. }
    \label{fig:rainrate}
\end{figure}
Similar to the approach from \cite{goodin}, we observe that the $SNR$ for a lidar is a function of both range and object reflectivity. Therefore it is more reliable to estimate minimum detectable power $P_{min}$ from max range which is often cited in manufacturer datasheets and then calculate $SNR$ for every point in the lidar scan individually. Therefore in our model, all points that have range-corrected power below $P_{min}$ are discarded since they cannot be measured reliably which is consistent with how lidars are built \cite{velo}. Further, small SNR contributes to large range uncertainty which we calculate through a model from the radar literature \cite{radar} as follows:
\begin{equation}
    \sigma_R = \frac{\Delta R}{\sqrt{2 \, SNR}},
    \label{eq:ran}
\end{equation}
where $\sigma_R$ is the standard deviation for range noise and $\Delta R$ is range accuracy determined by the finite bandwidth of the lidar system. Noisy range is calculated by adding normally distributed noise with zero mean and standard deviation of $\sigma_R$.

Another major problem with the existing lidar modelling algorithms is that they do not take beam divergence into account yet laser beams diffract during free space propagation. In other words, laser beam waist increases with propagation which reduces power available to the scatterers far away from the lidar sensor and hence affects PDF of randomly scattered points. We model range dependent waist size using Gaussian optics to keep the simulations simple although some lidars might use side emitting lasers with poor beam quality (non-Gaussian beam profiles). Width of a Gaussian beam can be calculated from
\begin{equation}
    w(R) = w_0 \sqrt{1 + \left( \frac{R}{R_0} \right)^2},
\end{equation}
where $w_0$ is the diffraction limited width and $R_0$ is the Rayleigh range \cite{sale}. For large propagation distances $R >> R_0$, the beam waist becomes
\begin{equation}
    w(R) \approx \frac{w_0}{R_0} R = \Theta R,
    \label{eq:gaus}
\end{equation}
where $\Theta$ is the divergence angle which can be found in lidar data sheets and is in the order of a few $mrad$.
\begin{algorithm} [H]
 \caption{mini-LISA without Monte-Carlo approach which is better suited for fog simulation} 
 \label{alg:minilisa}
 \algorithmicrequire{Lidar point cloud (pc)} \\
 \algorithmicensure{Noisy lidar point cloud (pcnew), label}
 \begin{algorithmic}
 \STATE {Init pcnew}
     \FOR{x,y,z,$\rho$(reflectivity) in pc}
         \STATE {Range $R = \sqrt{x^2 + y^2 + z^2}$}
         \STATE {Min power in arbitrary units $P_{min} = 0.9 R_{max}^{-2}$} 
         \STATE {Density of droplets $Nd = N(D,Rr)$}
         \STATE {Calculate extinction coefficient $\alpha$}
         \STATE {Back reflected power from the object in arbitrary units $P_0 = \frac{\rho e^{-2 \alpha R}}{R^2}$}
         \IF{$P_{0} < P_{min}$}
             \STATE {$R_{new} = 0$ and $\rho_{new} = 0$}
             \STATE {label = 0 for lost point}
         \ELSE
             \STATE {Original point with modified range and reflectivity:}
                 \STATE {Calculate range uncertainty $\sigma_R$}
                 \STATE {New range $R_{new} = R + \mathcal{N}(0,\sigma_R)$}
                 \STATE {New reflectivity $\rho_{new} = \rho e^{-2 \alpha R}$}
         \ENDIF
         \STATE {Calculate angles from old positions $\theta = cos^{-1}\left( z/R \right)$, $\phi = arctan2(y,x)$}
         \STATE {$x_{new} = R_{new} sin(\theta) cos(\phi)$}
         \STATE {$y_{new} = R_{new} sin(\theta) sin(\phi)$}
         \STATE {$z_{new} = R_{new} cos(\theta)$}
         \STATE {pcnew.append$ \left( x_{new}, y_{new}, z_{new}, \rho_{new} \right)$}
     \ENDFOR
 \end{algorithmic}
\end{algorithm}
Beam divergence leads to a quadratic scatterer range PDF. If we assume scatterers (i.e rain droplets) are distributed uniformly randomly in space, the number of particles illuminated by the lidar beam per range would be proportional to the beam cross-sectional area. Since beam width scales linearly in range as shown in Eq.~\ref{eq:gaus}, beam area scales quadratically. Therefore, in our Monte-Carlo simulations, random scatterer positions are selected from
\begin{equation}
    R_{rand} = R \, r^{-1/3},
    \label{eq:ran_pdf}
\end{equation}
where $R$ is the object range, $r$ is a random variable drawn uniformly from $[0, \, 1]$. A histogram of $R_{rand}$ should give rise to a quadratic profile. Since in our hybrid approach we are only considering large droplets for the Monte-Carlo part, their reflectivities are calculated using the Fresnel equation (at normal incidence) \cite{sale}
\begin{equation}
    \rho_{rand} = \left\lvert \frac{n - 1}{n + 1} \right\rvert^2,
\end{equation}
where $n$ is the complex refractive index of the scatterer. 

\begin{table}[h]
\centering
\caption{Percentage mean average precision of state of the art 3D detection networks on ``Car'' class in the KITTI dataset \cite{Geiger2013IJRR} for clear weather scenes and simulated rainy scenes using the algorithm from \cite{prin},\cite{goodin}, and LISA.}
\label{tab:1}
\resizebox{0.47\textwidth}{!}{%
\begin{tabular}{|c|c|ccc|}
\hline
  Network &   \begin{tabular}[c]{@{}c@{}} Weather  \end{tabular} & \multicolumn{3}{c|}{BEV / 3D} \\ \cline{3-5} 
 &  & Easy & Medium & Hard \\ \hline
 {PointPillars} & Clear & 89.98 / 83.08 & 86.52 / 74.83 & 84.19 / 72.77 \\ 
 & Bijelic \cite{prin} & 89.87 / 82.00 & 85.69 / 74.20 & 83.59 / 71.75 \\
 & Goodin \cite{goodin} & 31.33 / 18.76&20.73 / 13.56&18.00 / 11.66\\
 & LISA & 59.85 / 41.37 & 44.1 / 29.23 & 39.36 / 27.16 \\ \hline
  PV-RCNN & Clear & 90.14 / 89.30 & 88.05 / 79.40 & 87.54 / 78.81 \\
 & Bijelic & 90.04 / 88.92 & 87.99 / 79.16 & 87.29 / 78.60 \\
 & Goodin & 29.33 / 24.03 &20.67 / 18.20 &19.89 / 15.61 \\
 & LISA & 79.93 / 76.63 & 66.32 / 56.31 & 60.38 / 51.07 \\ \hline
  def PV-RCNN  & Clear & 90.14 / 89.27 & 87.89 / 82.99 & 87.55 / 78.76 \\
 & Bijelic & 90.09 / 89.15 & 87.95 / 82.36 & 87.48 / 78.64 \\
 & Goodin & 26.84 / 21.58  & 20.31 / 16.07 &16.70 / 15.18\\
 & LISA & 68.26 / 59.83 & 43.67 / 41.79 & 43.01 / 40.26 \\ \hline
\end{tabular}%
}
\end{table}

For large absorbing particles, Fresnel reflectivity is a good approximation to the backscattering efficiency \cite{bohren}. Finally the diameter of the random scatterers are selected from a given size distribution. For the Marshall-Palmer distribution, particle diameters are calculated from
\begin{equation}
    D_{rand} = -\frac{ln\left( 1 - r \right)}{\Lambda} + D_{st},
    \label{eq:diameter}
\end{equation}
where $D_{rand}$ is the random particle diameter, $r$ is a random variable drawn uniformly from $[0, \, 1]$, $\Lambda$ is the exponential decay constant as determined by the rain rate and $D_{st}$ is the size of the smallest particle that we would like to consider. A histogram of $D_{rand}$ should give rise to an exponentially decaying profile. Full lidar light scattering augmentation algorithm is presented in Algorithm~\ref{alg:lisa}. Marshall-Palmer distribution was used because it is fast to sample but any other particle size distribution would work here as well.

In addition to the rain augmentations, we augment lidar scenes in the KITTI dataset \cite{Geiger2013IJRR} with snow,  moderate advection fog, and strong advection fog. The algorithm for the simulation of snow is similar to that of rain. The simulation method for fog is described by algorithm \ref{alg:minilisa} which we call mini-LISA. Fog consists of dense small particles and therefore random scattering from large particles is ignored.
 The amount of snow in the scene is controlled by a precipitation rate in units of mm/hr. Fog simulations use a Gamma function for modelling particle distributions :
\begin{align}
    N(D) &= \frac{\gamma N_0 b^{(a+1)/\gamma}}{\Gamma\left( \frac{a+1}{\gamma}\right)} \left( \frac{D}{2} \right)^a e^{-b (D/2)^\gamma}, \\
    b &= \frac{a}{\gamma (D_C/2)^{\gamma}},
\end{align}
where $D$ is the particle diameter, $\Gamma$ is the Gamma function, $D_C$ is the mode (most probable) particle diameter, and $a$, $\gamma$ and $b$ are fitting parameters \cite{bmw}. We use parameters from \cite{bmw} for both fog and snow simulations. For fog simulations, the fitting parameters are fixed so the extinction coefficient is same across all lidar scans for the same fog type.

\section{Experiments and results}
In this section, we perform weather simulations on lidar point clouds of outdoor scenes from the KITTI dataset \cite{Geiger2013IJRR} and the Waymo Open Dataset \cite{waymo}. We compare recent state-of-the-art 3D object detection networks \cite{shi2020pv,Bhattacharyya2020DeformablePI,lang2019pointpillars,yan2018second,parta2} and analyze the effect of simulated weather on their performance.

\subsection{Simulation experiments}

Here, we describe the details of the rain simulation process. In order to perform simulations that reflect realistic adverse weather trends, we sample rain rates from an exponential distribution \cite{rainrate} at the rate $\lambda=0.05 \, mm/hr$, as can be seen in Fig.~\ref{fig:rainrate}. The maximum range of the lidar sensor ($R_{max}$) is chosen according to the specifications of the Velodyne sensor used for 3D data capture. In the case of \cite{Geiger2013IJRR}, we choose  $R_{max}=120 \, m$ \cite{datasheet}. Additional parameters that are used include  beam divergence of $B_{div}=3\times10^{-3}$ radians, and range accuracy of $\pm 4.5$ cm ($\Delta R=0.09 \, m$), and we consider particles with diameters of $D_{st} = 50 \mu m$ or above for random scattering. For the simulation method in \cite{prin}, we consider the Marshall-Palmer rain model \cite{bmw} with the parameter $\alpha$ corresponding to the values of rain rate used in the LISA simulation. 

\begin{figure*}[h]
	\begin{center}
	\includegraphics[width=.75\linewidth]{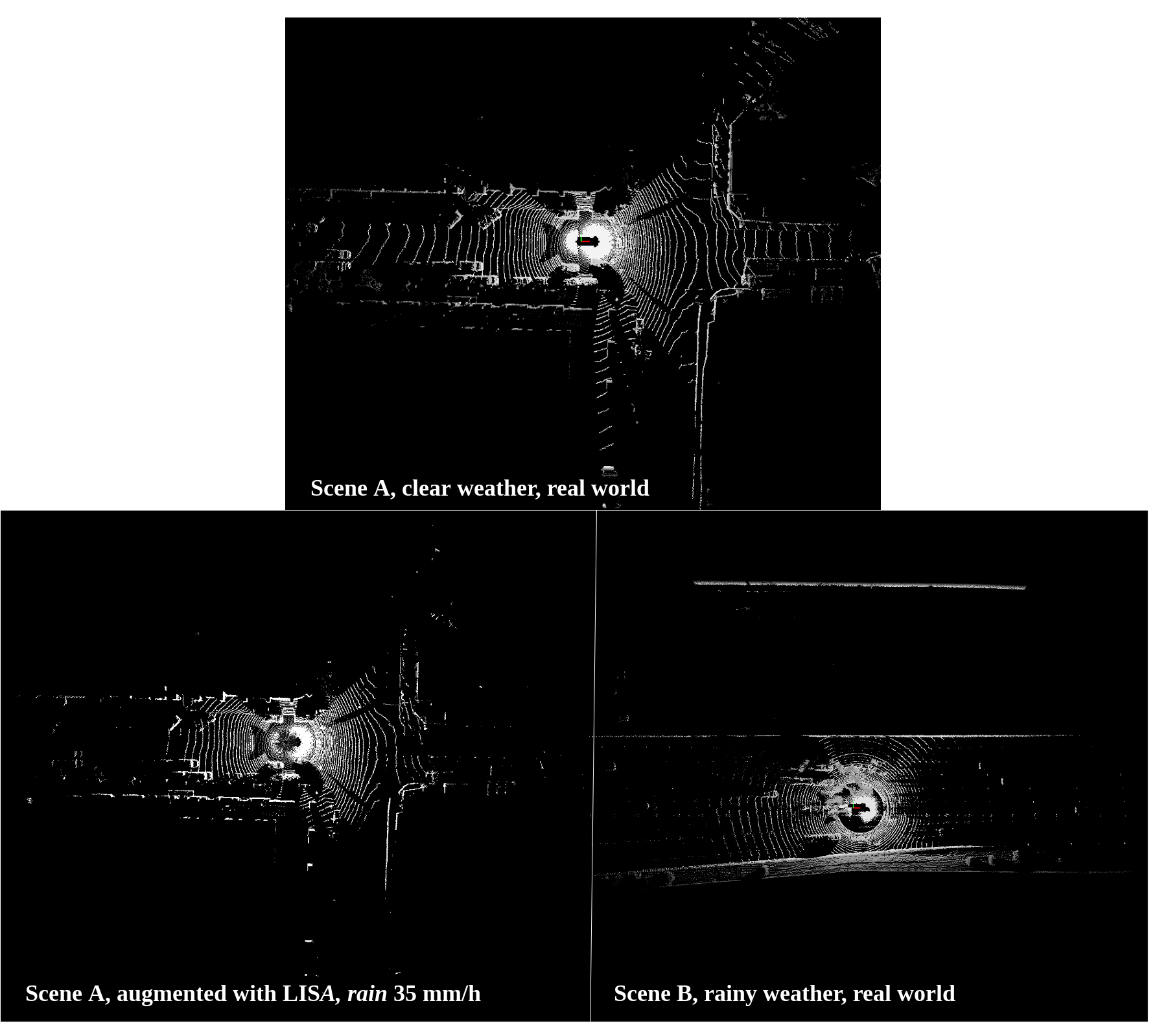}
\end{center}
	\caption{A qualitative comparison of the proposed augmentation method for simulating a rainy lidar scene against real world rainy data. Top: A clear weather scene (Scene A) from the Waymo Open Dataset \cite{waymo}. Bottom left: Scene A augmented using our method with rain rate 35 mm/hr. Bottom right: A real world rainy scene (Scene B) from \cite{waymo}. Rain leads to sparser scenes, increased range uncertainty and reduced visibility (best viewed zoomed in).}
	\label{fig:lidar_compare} 
\end{figure*}

Fig.~\ref{fig:lidar_compare} illustrates a scene from the Waymo Open Detection dataset \cite{waymo} under simulated rainy conditions. We compare these simulations against a real world rainy scene from the same dataset. As observed, our method results in the loss of points and reduced visibility similar to that of the scene with real rain.

\begin{figure*}[h]
    \centering
    \includegraphics[width=0.8\linewidth]{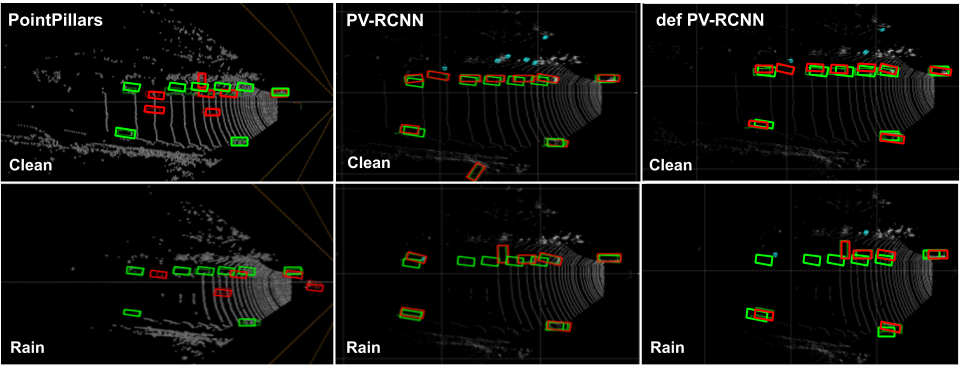}
    \caption{A comparison of detection before and after rain simulation for three networks trained on the KITTI dataset, PointPillars \cite{lang2019pointpillars} (first column), PV-RCNN \cite{shi2020pv} (second column), and deformable PV-RCNN \cite{Bhattacharyya2020DeformablePI} (third column). The first row depicts predictions of each network under normal conditions, while the second row depicts predictions under rainy conditions. The green bounding boxes are the ground truth annotations, while the red bounding boxes are the predictions. }
    \label{fig:comp_det}
\end{figure*}

\pagebreak

\subsection{Comparison of network performance improvements using simulated data}
In order to further compare the accuracy of the rain simulation methods, we retrain the networks evaluated in Table \ref{tab:2} with simulated data generated using the LISA, Bijelic \cite{prin}, and Goodin \cite{goodin} methods and evaluate them on real world rainy lidar scenes from \cite{waymo}. The comparisons may be seen in Table \ref{tab:retrain}.

The model from \cite{prin} is for fog and therefore does not take in rain rate to model rainy scenes. It does however take in an extinction coefficient. Here, we modify and in some sense improve that model by calculating the extinction coefficient from our average Mie treatment for a given rain rate. The extinction coefficient is then fed to their algorithm. In order to create rain/snow simulations on existing datasets, each sample is augmented with a different rain rate sampled from the distribution in Fig. \ref{fig:rainrate}. For evaluation and re-training, the test and training sets from each dataset are augmented and compared with the un-augmented samples and datasets of real-world adverse weather scenes. In most cases, retraining the networks \cite{yan2018second}, \cite{shi2020pv}, and \cite{parta2} on LISA simulated data results in the best performance on real world rainy scenes. This validates that a physics based modelling approach outperforms the existing approaches.

\begin{table*}[h]
\centering
\huge
\caption{A comparison of mean average precision values of state of the art networks retrained on simulated rainy lidar scenes using LISA, \cite{prin}, and \cite{goodin} evaluated on real world rainy scenes from the Waymo dataset \cite{Geiger2013IJRR} for class ``Vehicle''. The best performance for each category in in bold.}
\label{tab:retrain}
\resizebox{0.8\textwidth}{!}{%
\begin{tabular}{|c|c|cccc|}
\hline
\multicolumn{1}{|l|}{ } & \multicolumn{1}{l|}{ } & \multicolumn{4}{c|}{Real-world rain} \\ \cline{3-6} 
\multicolumn{1}{|l|}{Network} & \multicolumn{1}{l|}{Train/Test} & \multicolumn{4}{c|}{mAP/mAPH} \\ \cline{3-6} 
\multicolumn{1}{|l|}{} & \multicolumn{1}{l|}{} & Level 1 &  Improvement & Level 2 & Improvement \\ \hline
  SECOND \cite{yan2018second} & Clear  & 39.97 / 39.34 & -- & 39.40 / 38.96 & --\\
& LISA & \textbf{45.70} / \textbf{45.32} & \textit{5.73 / 5.98} &\textbf{45.23} / \textbf{44.86}& \textit{5.83 / 5.9} \\ 
 & Bijelic &42.12 / 41.34 & \textit{2.75 / 2.00} &41.71 / 40.94 & \textit{2.31 / 1.98}\\
 & Goodin &37.65 / 33.03 & \textit{-2.32 / -6.34} & 37.27 / 32.69 & \textit{-2.13 / -6.27}\\ \hline
   Part A\textasciicircum{2 \cite{parta2}} & Clear&48.51 / 47.83 & --  & 48.07 / 47.39 & --\\
 & LISA &\textbf{50.83} / \textbf{50.28}&\textit{2.32 / 2.45}& \textbf{50.33} / \textbf{49.79}&\textit{2.26 / 2.4}\\
 & Bijelic &44.03 / 43.31&\textit{-4.48 / -4.52} &43.64 / 42.93 & \textit{-4.43 / -4.46}\\
 & Goodin &34.51 / 33.16& \textit{-14.00 / -14.67} &34.17 / 32.83 &\textit{-13.90 / -14.56}\\ \hline
  PV-RCNN  \cite{shi2020pv}  & Clear& 42.17 / 41.25&--& 41.78 / 40.87 & --\\
& LISA & 44.95 / \textbf{43.72} & \textit{2.78 / 2.47}& 44.53 / \textbf{43.31} & \textit{2.75 / 2.44}\\
 & Bijelic &42.71 / 42.30&\textit{0.54 / 1.05} &42.30 / 41.89 &\textit{0.52 / 1.02}\\
 & Goodin & \textbf{46.67} / 43.34 &\textit{4.5 / 2.09}&\textbf{46.31} / 42.92&\textit{4.53 / 2.05}\\ \hline
\end{tabular}%
}
\end{table*}

\begin{table*}[h]
\centering
\caption{Percentage mean average precision of state of the art 3D detection networks on ``Vehicle'' class in the Waymo Open Dataset \cite{waymo} for clear weather scenes, real world rainy scenes, simulated rainy scenes using the algorithm from \cite{prin},\cite{goodin}, and LISA.}

\label{tab:2}
\resizebox{0.80\textwidth}{!}{
\hspace{0.1in}
\begin{tabular}{|c|c|cccc|}

\hline
  Network &   Weather & \multicolumn{4}{c|}{mAP/mAPH}  \\ \cline{3-6} 
&  & Level 1 & Degradation & Level 2 & Degradation\\ \hline
  SECOND & Clear & 58.37 / 57.69 &-- & 51.53 / 50.91  & --\\
 & Bijelic & 50.80 / 50.01& \textit{7.57 / 7.68} &44.03 / 43.34& \textit{7.5 / 7.57}\\
 & Goodin & 10.61 / 10.36& \textit{47.76 / 47.33}&9.21 / 8.98& \textit{42.32 / 41.93}\\
 & LISA & 28.87 / 28.39& \textit{29.5 / 29.3} & 24.86 / 24.44& \textit{26.67 / 26.47}\\
 & Real world rain & 39.97 / 39.34& \textit{18.4 / 18.25} & 39.40 / 38.96& \textit{12.13 / 11.95}\\ \hline
  Part A\textasciicircum{}2 & Clear &64.09 / 63.30&-- &55.83 / 55.15& --\\
 & Bijelic & 58.80 / 57.88& \textit{5.29 / 5.42} & 50.89 / 50.10& \textit{4.94 / 5.05}\\
 & Goodin & 11.94 / 11.71& \textit{52.51 / 51.95}&10.47 / 10.25& \textit{45.36 / 44.9}\\
 & LISA & 31.76 / 31.15& \textit{32.33 / 32.15} & 27.33 / 26.81& \textit{28.50 / 28.34}\\
 & Real world rain & 48.51 / 47.83& \textit{15.58 / 15.47} & 48.07 / 47.39& \textit{7.76 / 7.76}\\ \hline
  PV-RCNN & Clear & 63.88 / 62.93 & --&55.58  / 54.76&--\\
 & Bijelic & 57.38 / 56.32 & \textit{6.5 / 6.61} &49.66 / 48.75& \textit{5.92 / 6.01}\\
 & Goodin &11.74 / 11.07& \textit{52.14 / 51.86} &10.33 / 9.71& \textit{45.26 / 45.05}\\
 & LISA & 35.42 / 34.52& \textit{28.43 / 28.41} &30.67 / 29.88& \textit{24.91 / 24.88}\\
 & Real world rain &42.17 / 41.25& \textit{21.71 / 21.68} &41.78 / 40.87& \textit{13.8 / 13.89}\\ \hline
\end{tabular}%
}
\end{table*}

\subsection{Analyzing the effect of rain on 3D object detection}

In this section, we analyze the effect of simulated rain on recent state-of-the-art detection approaches. We conduct two sets of experiments. (i) We train  PointPillars \cite{lang2019pointpillars}, 
PV-RCNN \cite{shi2020pv}, and Deformable PV-RCNN \cite{Bhattacharyya2020DeformablePI} on the KITTI lidar dataset \cite{Geiger2013IJRR} and evaluate them on simulated rainy lidar data generated using Bijelic \cite{prin}, Goodin \cite{goodin}, and the proposed simulation method. The results of these evaluations can be seen in Table \ref{tab:1}. (ii) In order to have a comparison with real world adverse weather data, we train SECOND \cite{yan2018second}, PV-RCNN \cite{shi2020pv}, and PartA\textasciicircum2 \cite{parta2} on the Waymo Open Dataset \cite{waymo} and evaluate them on the simulated scenes as well as real world rainy scenes from \cite{waymo}. The results of these evaluations can be seen in Table \ref{tab:2}.  We make the following observations:
\begin{itemize}[topsep=0pt,noitemsep,leftmargin=*]
    \item The performance of each network across both datasets drops when evaluated on the simulated rainy lidar scenes. This is due to the presence of noise and the loss of information by way of lost points in the scene.
    
    \item Detection networks are affected most negatively by the occurrence of missing points from the scene caused by the laser intensity attenuation and random points back-scattering more power than the target. Capturing lidar data in adverse weather can result in the deletion of entire objects from the scene, especially those at a large distance away from the sensor. In Table \ref{tab:1}, this is reflected in the degraded performance of the networks in the ``Medium" and ``Hard" categories on the simulated data from our method for networks trained on KITTI data.   
    
    \item PointPillars, since it encodes entire columns, is particularly prone to the problem of missing data points and has performance most affected by rain. The loss of points can result in the loss of entire voxels/pillars as well as negatively affect the quality of the voxel features extracted by the encoder. Point based methods, and those that utilize a combination of voxelization and set abstraction, are less prone to this loss in information in the feature space as indicated by the relative improved performance of the other networks compared to that of PointPillars.
    
    \item Noticing the disparity in performance on LISA data and that of Bijelic \etal and Goodin \etal, we draw the conclusion that the missing data points contribute the most to missed detections. The method from \cite{goodin} overestimates the effect of rain and results in scenes with too many missing and scattered points. This is reflected in the extremely low mAP values in both Table \ref{tab:1} and \ref{tab:2}. The method from \cite{prin} underestimates the effect of rain on lidar scenes, as can be seen by the negligible drop in performance in Table \ref{tab:1}. 

    \item  Comparing the predictions of PointPillars, PV-RCNN, and deformable PV-RCNN on lidar scenes from \cite{Geiger2013IJRR} under normal and moderate rain conditions (rain rate 10 mm/hr), Fig.~\ref{fig:comp_det}  demonstrates visually how networks maybe be affected by the simulated adverse weather. PV-RCNN and deformable PV-RCNN provide more accurate predictions (red bounding boxes) that line up with the ground truth annotations (green bounding boxes) than PointPillars, with deformable PV-RCNN having fewer false predictions than PV-RCNN. However, all networks are adversely affected by rain, with false detection around scattered points as well as several missed detections due to the loss of points.
    
\end{itemize}

\subsection{Fog and snow}

\begin{figure*}[ht]
    \centering
    \includegraphics[width=0.8\textwidth]{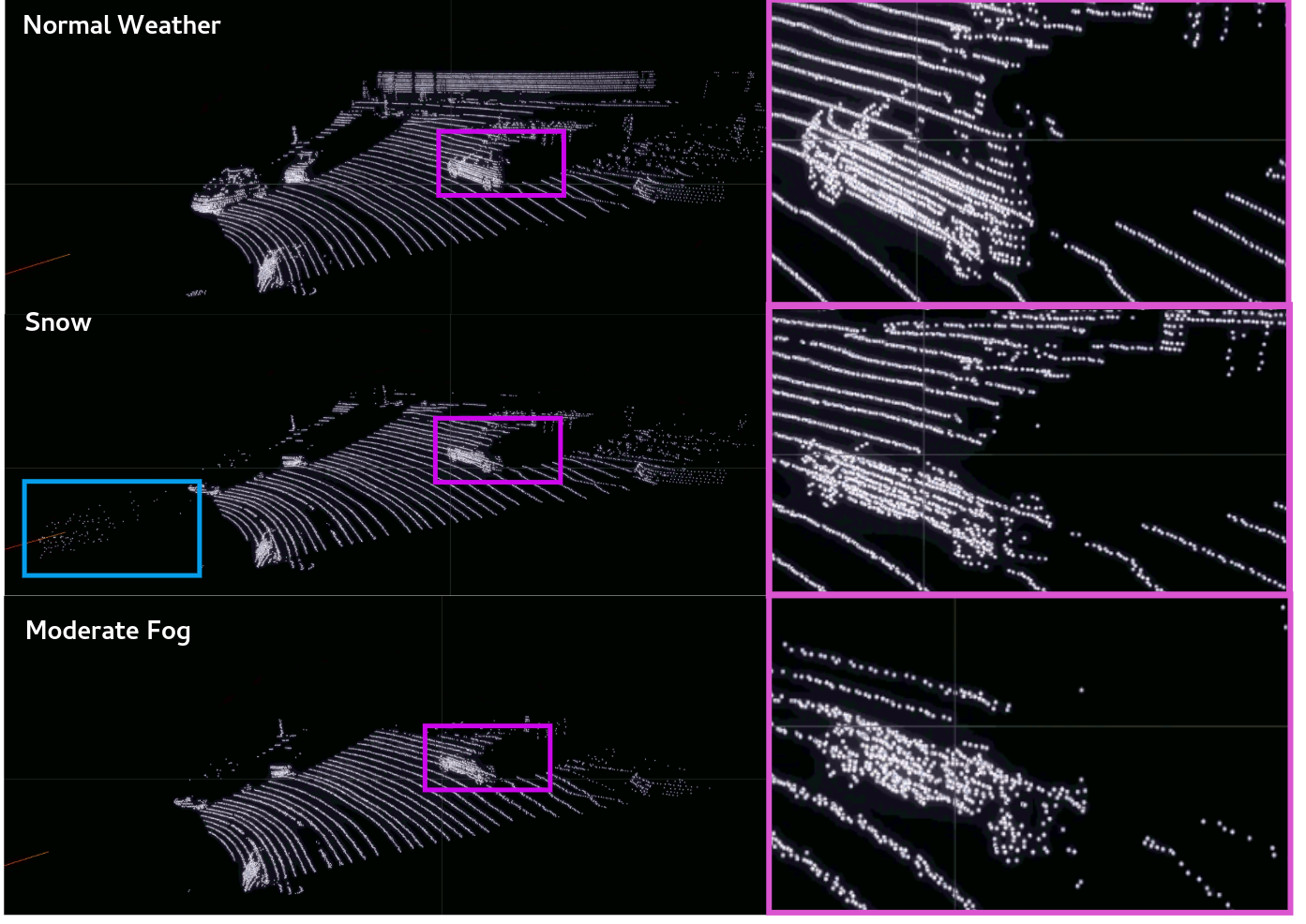}
    \caption{A visual comparison of snow and fog simulations on the KITTI dataset. The top image is a scene in normal weather conditions. The middle image shows a lidar scene in simulated snow conditions using the LISA method. The scattered points near the sensor are bounded in blue. The last row shows the same scene in simulated moderate fog conditions. A closer view of a car can be seen to the right of each image.}
    \label{fig:viz}
\end{figure*}

The visual effect of simulating snow and fog on lidar scenes from KITTI can be seen in Figure \ref{fig:viz}. The effect of snow is similar to that of rain, with lost points farther away from the sensor and the presence of scattered points near the sensor. This is indicated by a blue bounding box in the middle image. With the presence of snow, the amount of scattering is higher than that of a rainy scene as expected (this is why snow appears white). The nature of missing points also differs from snow and fog scenes. The presence of fog greatly reduces visibility, and all points beyond a certain distance from the sensor are lost, more so than in the case of snow. This can be discerned from comparing the middle and last images in the first column. The presence of fog also induces a ``fuzzy" effect on the points, causing objects to further lose their structural fidelity due to increased range uncertainty. 

\begin{table}[h]
\centering
\huge
\caption{A comparison of mean average precision values of state of the art networks for the ``Car" class of the KITTI dataset \cite{Geiger2013IJRR} on simulated fog and snow conditions.}
\label{tab:1}
\resizebox{\linewidth}{!}{%
\begin{tabular}{|c|c|ccc|}
\hline  
  Network &   Simulation  & \multicolumn{3}{c|}{BEV/3D} \\ \cline{3-5} 
 &  & Easy & Medium & Hard \\ \hline
  Pointpillars & Clear & \multicolumn{1}{l}{ 89.98 / 83.08} & \multicolumn{1}{l}{86.52 / 74.83} & \multicolumn{1}{l|}{84.19 / 72.77} \\
 & Snow & 63.05 / 43.96 & 46.41 / 30.27 & 40.95 / 28.53 \\
 & \begin{tabular}[c]{@{}c@{}}Mod. Fog\end{tabular} & 55.10 / 43.12 & 36.48 / 28.71 & 35.68 / 26.60 \\
 & \begin{tabular}[c]{@{}c@{}}Strong Fog\end{tabular} & 45.24 / 33.35 & 29.54 / 23.23 & 27.68 / 20.60 \\ \hline
  PV-RCNN & Clear & \multicolumn{1}{l}{90.14 / 89.30} & \multicolumn{1}{l}{ 88.05 / 79.40} & \multicolumn{1}{l|}{87.54 / 78.81} \\
 & Snow & 79.76 / 76.62 & 60.80 / 56.32 & 60.10 / 51.10 \\
 & \begin{tabular}[c]{@{}c@{}}Mod. Fog\end{tabular} & 69.68/64.62 & 48.36/42.96 & 43.93/41.00 \\
 & \begin{tabular}[c]{@{}c@{}}Strong Fog\end{tabular} & 57.50/54.59 & 39.42/31.17 & 35.44/34.06 \\ \hline
\multicolumn{1}{|l|}{  def PV-RCNN} & Clear & \multicolumn{1}{l}{90.14 / 89.27} & \multicolumn{1}{l}{87.89 / 82.99} & \multicolumn{1}{l|}{87.55 / 78.76} \\
\multicolumn{1}{|l|}{} & Snow & 79.11 / 73.71 & 60.24 / 55.50 & 59.99 / 50.53 \\
\multicolumn{1}{|l|}{} & \begin{tabular}[c]{@{}c@{}}Mod. Fog\end{tabular} & 72.83/66.30 & 48.14 / 43.55 & 44.26 / 41.88 \\
\multicolumn{1}{|l|}{} & \begin{tabular}[c]{@{}c@{}}Strong Fog\end{tabular} & 58.93 / 54.81 & 39.22 / 35.48 & 34.68 / 34.68 \\ \hline
\end{tabular}%
}
\end{table}

\subsection{Future Work}
A rigorous comparison between the augmentation models requires measurements under real rain as opposed to artificial rain from an environment chamber with corresponding ground truth clear scans (i.e with the same objects in the same position). Collecting such data is challenging and currently there is no such dataset publicly available. However, from physical arguments, we claim LISA is more accurate than the other algorithms in the literature. Experimental verification of these claims remains a future study direction. Modified/improved model from Bijelic \etal as well as LISA and the model from Goodin \etal will be made publicly available to facilitate research in this direction.

\pagebreak

\section{Conclusions}
In summary, we present an efficient physics-based simulation method for augmenting lidar data taken under clear weather conditions to make 3D object detectors robust against noise introduced by adverse weather conditions. The simulation method relies on a hybrid Monte Carlo approach where the effects of small scatterers are treated on average. Effects of large particles are calculated by randomly placing them in the laser beam path and calculating the back-scattered power. Preliminary results on the real world rainy scenes from the Waymo dataset shows significant improvements in mAP values. Furthermore, we analyzed the effect of weather on recent state-of-the-art lidar-based object detectors and proposed promising future avenues for developing weather-robust object detectors. 

\pagebreak

{\small
\bibliographystyle{ieee_fullname}
\bibliography{egbib}

\begin{thebibliography}{10}\itemsep=-1pt

\bibitem{Bhattacharyya2020DeformablePI}
P. Bhattacharyya and K. Czarnecki.
\newblock Deformable pv-rcnn: Improving 3d object detection with learned
  deformations.
\newblock {\em ArXiv}, abs/2008.08766, 2020.

\bibitem{prin}
M. Bijelic, T. Gruber, F. Mannan, F. Kraus, W. Ritter, K. Dietmayer, and F.
  Heide.
\newblock Seeing through fog without seeing fog: Deep multimodal sensor fusion
  in unseen adverse weather.
\newblock In {\em The IEEE Conference on Computer Vision and Pattern
  Recognition (CVPR)}, June 2020.

\bibitem{bohren}
C.~F. Bohren and D.~R. Huffman.
\newblock {\em Absorption and Scattering of Light by Small Particles}.
\newblock Wiley, 1998.

\bibitem{canny}
J. {Canny}.
\newblock A computational approach to edge detection.
\newblock {\em IEEE Transactions on Pattern Analysis and Machine Intelligence},
  PAMI-8(6):679--698, 1986.

\bibitem{chen2017multi}
Xiaozhi Chen, Huimin Ma, Ji Wan, Bo Li, and Tian Xia.
\newblock Multi-view 3d object detection network for autonomous driving.
\newblock In {\em Proceedings of the IEEE Conference on Computer Vision and
  Pattern Recognition}, pages 1907--1915, 2017.

\bibitem{radar}
G.~R. Curry.
\newblock {\em Radar System Performance Modeling}.
\newblock Artech House, 2 edition, 2005.

\bibitem{Geiger2013IJRR}
Andreas Geiger, Philip Lenz, Christoph Stiller, and Raquel Urtasun.
\newblock Vision meets robotics: The kitti dataset.
\newblock {\em International Journal of Robotics Research (IJRR)}, 2013.

\bibitem{goodin}
C. Goodin, D. Carruth, M. Doude, and C. Hudson.
\newblock Predicting the influence of rain on lidar in adas.
\newblock {\em Electronics}, 8, 2019.

\bibitem{velo}
D.~S. Hall.
\newblock High definition lidar system.
\newblock Velodyne Lidar Inc, Patent US7969558B2.

\bibitem{he2020structure}
Chenhang He, Hui Zeng, Jianqiang Huang, Xian-Sheng Hua, and Lei Zhang.
\newblock Structure aware single-stage 3d object detection from point cloud.
\newblock In {\em Proceedings of the IEEE/CVF Conference on Computer Vision and
  Pattern Recognition}, pages 11873--11882, 2020.

\bibitem{Heinzler}
R. {Heinzler}, F. {Piewak}, P. {Schindler}, and W. {Stork}.
\newblock Cnn-based lidar point cloud de-noising in adverse weather.
\newblock {\em IEEE Robotics and Automation Letters}, 5(2):2514--2521, 2020.

\bibitem{ku2019monocular}
Jason Ku, Alex~D Pon, and Steven~L Waslander.
\newblock Monocular 3d object detection leveraging accurate proposals and shape
  reconstruction.
\newblock In {\em Proceedings of the IEEE Conference on Computer Vision and
  Pattern Recognition}, pages 11867--11876, 2019.

\bibitem{lang2019pointpillars}
Alex~H Lang, Sourabh Vora, Holger Caesar, Lubing Zhou, Jiong Yang, and Oscar
  Beijbom.
\newblock Pointpillars: Fast encoders for object detection from point clouds.
\newblock In {\em Proceedings of the IEEE Conference on Computer Vision and
  Pattern Recognition}, pages 12697--12705, 2019.

\bibitem{liang2019multi}
Ming Liang, Bin Yang, Yun Chen, Rui Hu, and Raquel Urtasun.
\newblock Multi-task multi-sensor fusion for 3d object detection.
\newblock In {\em Proceedings of the IEEE Conference on Computer Vision and
  Pattern Recognition}, pages 7345--7353, 2019.

\bibitem{datasheet}
Velodyne Lidar.
\newblock \url{https://velodynelidar.com/products/hdl-64e/#downloads}.

\bibitem{rainrate}
Dmitri Moisseev and V Chandrasekar.
\newblock Examination of the $\mu$ - $\lambda$ relation suggested for drop size
  distribution parameters.
\newblock {\em Journal of Atmospheric and Oceanic Technology}, 24, 06 2007.

\bibitem{pang2020clocs}
Su Pang, Daniel Morris, and Hayder Radha.
\newblock Clocs: Camera-lidar object candidates fusion for 3d object detection.
\newblock {\em arXiv preprint arXiv:2009.00784}, 2020.

\bibitem{qi2018frustum}
Charles~R Qi, Wei Liu, Chenxia Wu, Hao Su, and Leonidas~J Guibas.
\newblock Frustum pointnets for 3d object detection from rgb-d data.
\newblock In {\em Proceedings of the IEEE conference on computer vision and
  pattern recognition}, pages 918--927, 2018.

\bibitem{REF:qi2017pointnet}
Charles~R Qi, Hao Su, Kaichun Mo, and Leonidas~J Guibas.
\newblock Pointnet: Deep learning on point sets for 3d classification and
  segmentation.
\newblock {\em Proc. Computer Vision and Pattern Recognition (CVPR), IEEE},
  2017.

\bibitem{REF:qi2017pointnetplusplus}
Charles~R Qi, Li Yi, Hao Su, and Leonidas~J Guibas.
\newblock Pointnet++: Deep hierarchical feature learning on point sets in a
  metric space.
\newblock {\em arXiv preprint arXiv:1706.02413}, 2017.

\bibitem{bmw}
R.~H. Rasshofer, M. Spies, and H. Spies.
\newblock Influences of weather phenomena on automotive laser radar systems.
\newblock {\em Advances in Radio Science}, 9:49--60, 2011.

\bibitem{sale}
B.~E.~A. Saleh and M.~C. Teich.
\newblock {\em Fundamentals of Photonics}.
\newblock John Wiley and Sons, Inc., 2 edition, 1991.

\bibitem{shi2020pv}
Shaoshuai Shi, Chaoxu Guo, Li Jiang, Zhe Wang, Jianping Shi, Xiaogang Wang, and
  Hongsheng Li.
\newblock Pv-rcnn: Point-voxel feature set abstraction for 3d object detection.
\newblock In {\em Proceedings of the IEEE/CVF Conference on Computer Vision and
  Pattern Recognition}, pages 10529--10538, 2020.

\bibitem{shi2019pointrcnn}
Shaoshuai Shi, Xiaogang Wang, and Hongsheng Li.
\newblock Pointrcnn: 3d object proposal generation and detection from point
  cloud.
\newblock In {\em Proceedings of the IEEE Conference on Computer Vision and
  Pattern Recognition}, pages 770--779, 2019.

\bibitem{Shi2019PartA2N3}
Shaoshuai Shi, Zhe Wang, X. Wang, and Hongsheng Li.
\newblock Part-a2 net: 3d part-aware and aggregation neural network for object
  detection from point cloud.
\newblock {\em ArXiv}, abs/1907.03670, 2019.

\bibitem{parta2}
Shaoshuai Shi, Zhe Wang, Xiaogang Wang, and Hongsheng Li.
\newblock Part-a\({}^{\mbox{2}}\) net: 3d part-aware and aggregation neural
  network for object detection from point cloud.
\newblock {\em CoRR}, abs/1907.03670, 2019.

\bibitem{pymie}
B.~J. Sumlin.
\newblock \url{https://github.com/bsumlin/PyMieScatt}.

\bibitem{waymo}
Pei Sun, Henrik Kretzschmar, Xerxes Dotiwalla, Aurelien Chouard, Vijaysai
  Patnaik, Paul Tsui, James Guo, Yin Zhou, Yuning Chai, Benjamin Caine, et~al.
\newblock Scalability in perception for autonomous driving: Waymo open dataset.
\newblock In {\em Proceedings of the IEEE/CVF Conference on Computer Vision and
  Pattern Recognition}, pages 2446--2454, 2020.

\bibitem{ulbrich}
C.~W. Ulbrich and D. Atlas.
\newblock {Extinction of Visible and Infrared Radiation in Rain: Comparison of
  Theory and Experiment}.
\newblock {\em Journal of Atmospheric and Oceanic Technology}, 2(3):331--339,
  09 1985.

\bibitem{yan2018second}
Yan Yan, Yuxing Mao, and Bo Li.
\newblock Second: Sparsely embedded convolutional detection.
\newblock {\em Sensors}, 18(10):3337, 2018.

\bibitem{yang2018pixor}
Bin Yang, Wenjie Luo, and Raquel Urtasun.
\newblock Pixor: Real-time 3d object detection from point clouds.
\newblock In {\em Proceedings of the IEEE Conference on Computer Vision and
  Pattern Recognition}, pages 7652--7660, 2018.

\bibitem{yang20203dssd}
Zetong Yang, Yanan Sun, Shu Liu, and Jiaya Jia.
\newblock 3dssd: Point-based 3d single stage object detector.
\newblock In {\em Proceedings of the IEEE/CVF Conference on Computer Vision and
  Pattern Recognition}, pages 11040--11048, 2020.

\bibitem{ipod}
Zetong Yang, Yanan Sun, Shu Liu, Xiaoyong Shen, and Jiaya Jia.
\newblock Ipod: Intensive point-based object detector for point cloud.
\newblock 12 2018.

\bibitem{yang2019std}
Zetong Yang, Yanan Sun, Shu Liu, Xiaoyong Shen, and Jiaya Jia.
\newblock Std: Sparse-to-dense 3d object detector for point cloud.
\newblock In {\em Proceedings of the IEEE International Conference on Computer
  Vision}, pages 1951--1960, 2019.

\bibitem{zhou2018voxelnet}
Yin Zhou and Oncel Tuzel.
\newblock Voxelnet: End-to-end learning for point cloud based 3d object
  detection.
\newblock In {\em Proceedings of the IEEE Conference on Computer Vision and
  Pattern Recognition}, pages 4490--4499, 2018.

\bibitem{REF:zhou2017voxelnet}
Yin Zhou and Oncel Tuzel.
\newblock Voxelnet: End-to-end learning for point cloud based 3d object
  detection.
\newblock 2018.

\end{thebibliography}
}

\end{document}